\documentclass[10pt,twocolumn,letterpaper,pagebackref,breaklinks,colorlinks,allcolors=cvprblue]{article}
 
\usepackage[pagenumbers]{cvpr} 

\usepackage{booktabs}
\usepackage{orcidlink}
\usepackage{graphicx} 
\usepackage{multirow} %
\usepackage[accsupp]{axessibility}  

\definecolor{cvprblue}{rgb}{0.21,0.49,0.74}

\def\paperID{2} 
\def\confName{MSLR}
\def\confYear{2026}

\title{Sign Language Recognition in the Age of LLMs}

\author{
Václav Javorek$^{1,2}$ \quad
Jakub Honzík$^{1}$ \quad
Ivan Gruber$^{1}$ \quad
Tomáš Železný$^{1}$ \quad
Marek Hrúz$^{1}$\\
$^{1}$University of West Bohemia, Czech Republic \quad
$^{2}$Eindhoven University of Technology, The Netherlands\\
{\tt\small v.javorek@tue.nl \quad honzikj@fav.zcu.cz \quad grubiv@fav.zcu.cz}\\
{\tt\small zeleznyt@fav.zcu.cz \quad mhruz@fav.zcu.cz}
}

\begin{document}
\maketitle              

\begin{abstract}
Recent Vision Language Models (VLMs) have demonstrated strong performance across a wide range of multimodal reasoning tasks. This raises the question of whether such general-purpose models can also address specialized visual recognition problems such as isolated sign language recognition (ISLR) without task-specific training. In this work, we investigate the capability of modern VLMs to perform ISLR in a zero-shot setting. We evaluate several open-source and proprietary VLMs on the WLASL300 benchmark. Our experiments show that, under prompt-only zero-shot inference, current open-source VLMs remain far behind classic supervised ISLR classifiers by a wide margin. However, follow-up experiments reveal that these models capture partial visual–semantic alignment between signs and text descriptions. Larger proprietary models achieve substantially higher accuracy, highlighting the importance of model scale and training data diversity. All our code is publicly available on GitHub.\footnote{\url{https://github.com/VaJavorek/WLASL_LLM}}.
\end{abstract}

\section{Introduction}

\begin{figure}[!ht]
    \centering
    \includegraphics[width=0.8\linewidth]{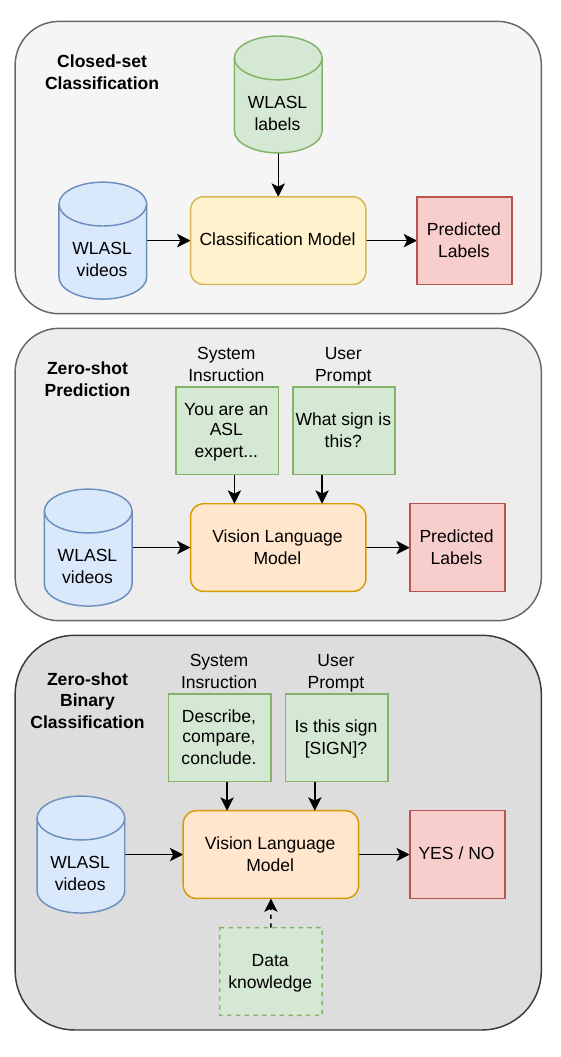}
    \caption{Evaluation paradigms for zero-shot ISLR with VLMs. \textbf{Top:} Standard closed-set classification with defined target labels. \textbf{Middle:} Zero-shot prediction instructing the model to behave as an ASL expert capable of categorizing the input videos into an open set. \textbf{Bottom:} Zero-shot binary classification instructing the model to think about the input video to conclude whether the performed sign is the one the user thinks of, with optional knowledge about the data.}
    \label{fig:diagram}
\end{figure}

Sign languages (SL)s are fully expressive, natural languages that rely on visual-manual modalities, including handshapes, movements, facial expressions, and body posture. Understanding sign languages is critical for accessibility, inclusion, and human-computer interaction, yet automated recognition remains challenging due to the high-dimensional spatio-temporal complexity and subtle linguistic structure of the signs. Although individual signs are an important part of SL (similar to vocabulary), co-articulation tends to be more complex and "wild" (similar to speech). In this regard we can separate the works on automated SL understanding to Isolated Sign Language Recognition (ISLR) and Sign Language Translation (SLT). Recent studies are involved in a more complex problem of (gloss-free) SLT, but tend to forget about ISLR. The models learn relationships between SL and written language, but seldom show a well-grounded behavior in the form of understanding isolated signs without the context. Also the interpretability of the translation models is far behind the onslaught of the newly designed modules modeling niche attributes of the spatio-temporal essence of the SLs. In this work, we want to take a step back to observe how the most recent Large Language Models (LLM)s and Vision Language Models (VLM)s handle the seemingly easier task of ISLR.

Traditionally, ISLR systems have relied on task-specific supervised learning, requiring large annotated datasets and specialized architectures to model hand motion, body pose, and temporal dynamics. While recent deep learning methods, including VAP~\cite{VAP-10.1007/978-3-031-72946-1_20}, LLaVA-SLT~\cite{liang2024llava}, Sign2GPT~\cite{wong2024signgpt}, SSLT~\cite{NEURIPS2024_ced76a66}, SSVP-SLT~\cite{rust2024towards}, Uni-Sign~\cite{li2025unisign}, Geo-Sign~\cite{fish2025geosign}, SHuBERT~\cite{gueuwou-etal-2025-shubert}, SignMusketeers~\cite{gueuwou-etal-2025-signmusketeers}, LITFIE~\cite{2025arXiv251208040J}, have achieved impressive performance on benchmark datasets, these models are limited by their dependence on labeled data and struggle to generalize across signers, environments, and languages. Even the benchmark datasets are often limited to a certain domain of expressivity and fall short of everyday SL usage.
At the same time, LLMs and VLMs~\cite{qwen3-vl, gpt-5, gemini2.5, llavanextvideo, internvl3.5, bagel, nvidia2025nvidianemotronnanov2} have demonstrated remarkable capabilities in zero-shot and few-shot reasoning, including tasks involving multimodal input such as images and video. These models leverage massive pretraining and powerful sequence modeling to generate coherent textual outputs from visual observations, suggesting a potential avenue for tackling sign language recognition without task-specific training. However, their application to SLs remains largely unexplored. 

This work investigates whether VLMs can serve as zero-training ISLR models. Specifically, we explore the ability of state-of-the-art VLMs to recognize individual signs using only prompt-based visual input, without fine-tuning. By framing ISLR as a multimodal reasoning problem, we aim to assess the emergent capability of VLMs to perceive and interpret structured visual language, opening a new direction for SLT research that goes beyond conventional supervised pipelines.

Our contributions are threefold:
\begin{itemize}
    \item We present the first systematic study of zero-shot ISLR with VLMs, evaluating current off-the-shelf capabilities without any ISLR-specific training, and find that while individual sign recognition can work surprisingly well in this setting, state-of-the-art performance still typically requires fine-tuning.
    \item We analyze the impact of different prompt designs and visual input strategies on model performance, highlighting the limitations and strengths of current VLM architectures for fine-grained visual linguistic understanding.
    \item We provide insights into the potential of VLMs as general-purpose visual-linguistic learners, offering a fresh perspective on how future multimodal models could bridge the gap between language and perception in complex structured domains.
\end{itemize}
 
\section{Related work}

\subsection{Visual Language Modeling}
Recent advances in video understanding and video-to-sequence modeling have been largely driven by the development of VLMs, which integrate pretrained vision encoders with large language models to enable joint visual–textual reasoning. Early works such as Flamingo~\cite{NEURIPS2022_960a172b} introduced a cross-attention-based architecture, leveraging a Perceiver-like module to efficiently handle variable-length visual inputs and enable few-shot or zero-shot multimodal reasoning. Subsequent models such as BLIP-2~\cite{pmlr-v202-li23q} also use cross-attention to align visual embeddings with LLMs, while models like LLaVA~\cite{NEURIPS2023_6dcf277e} employ lightweight projection layers to map visual features into the LLM embedding space. These approaches demonstrate that large language models can be adapted to visual domains without full end-to-end training, enabling flexible and generalizable multimodal reasoning.

Recent VLMs have extended these ideas to video understanding and multimodal reasoning. Models such as LLaVA-NeXT~\cite{llavanextvideo}, InternVL~\cite{internvl3.5}, and the Qwen-VL~\cite{qwen3-vl} family encode temporal visual inputs using pretrained vision backbones and aggregate features for LLM processing. Similarly, frontier multimodal models like GPT-5~\cite{gpt-5} and Gemini~\cite{gemini2.5} support high-resolution images and videos and demonstrate strong zero-shot and few-shot performance. These models are typically applied to tasks such as visual question answering, image and video captioning, action recognition, and instruction-guided reasoning, leveraging frozen LLMs with lightweight adapters or projection layers to generalize without full end-to-end training.

Despite their scale and generality, existing VLMs are primarily evaluated on natural image and video datasets and are not designed for sign language. Fine-grained hand articulation, body motion, and linguistically structured visual patterns in sign language remain largely underexplored. In particular, the capability of these models to perform isolated sign language recognition without task-specific training has received little attention, motivating this work.

\subsection{Sign Language Recognition}







The sign language recognition field covers several tasks that differ in the structure of the input video and the type of predicted output. ISLR systems process short video clips containing a single sign and predicts the corresponding gloss label. Continuous Sign Language Recognition (CSLR) processes unsegmented videos with multiple consecutive signs, requiring the model to detect sign boundaries and predict a sequence of glosses. Sign Language Translation (SLT) extends this further by translating a continuous sign video directly into a spoken language sentence, combining visual recognition with semantic and linguistic understanding either by first producing a sequence of glosses or directly outputting words of the target language.

While SLT is the most challenging of these tasks and has recently received a lot of attention~\cite{NEURIPS2024_ced76a66,rust2024towards,li2025unisign,fish2025geosign,2025arXiv251208040J}, we specifically focus on ISLR in this work. ISLR provides a controlled setting where each video contains a single sign, making it suitable for evaluating whether large language models can recognize signs without task-specific training. By focusing on individual signs, we can study the zero-shot recognition capabilities of VLMs and assess their potential for sign language understanding in more complex tasks.

Recent ISLR methods employ various architectures to capture the spatiotemporal dynamics of sign gestures. ST‑GCN~\cite{Yan_Xiong_Lin_2018} applies spatial‑temporal graph convolutions over pose keypoints to explicitly capture the motion and interaction patterns of hands and body joints, serving as a strong skeleton‑based baseline. SignBERT~\cite{Hu_2021_ICCV} introduces self‑supervised pre‑training by treating pose sequences as masked tokens for contextual reconstruction, and SignBERT+~\cite{10109128} incorporates model‑aware hand priors and multi‑level masking to better capture hierarchical pose context. MSLU~\cite{DBLP_journals_corr_abs-2408-08544} emphasizes multi‑stream learning by jointly aligning pose and appearance cues with unified objectives, helping models exploit complementary modalities. Larger unified frameworks such as Uni‑Sign~\cite{li2025unisign} propose large‑scale generative pre‑training with prior‑guided fusion of pose and RGB features to close the gap between pre‑training and downstream tasks.

\subsection{Prompt Design Strategies}

Carefully designed inputs can steer LLMs toward new tasks without updating their parameters. In multimodal settings, prompts may include textual instructions, visual examples, or structured templates that guide the model’s reasoning process.
Brown et al.~\cite{brown2020language} introduce few-shot prompting, demonstrating that language model behavior can be steered through prompt design without task-specific fine-tuning. Wei et al.~\cite{wei2022chain} show that instructing models to generate intermediate reasoning steps, referred to as chain-of-thought prompting, significantly improves performance on complex reasoning tasks. Wang et al.~\cite{wang2022self} extend this idea with self-consistency decoding, where multiple reasoning paths are sampled and aggregated to improve robustness.
Beyond manual prompt design, several approaches learn prompts automatically. Li et al.~\cite{li2021prefix} introduce training of continuous prompts while keeping the language model frozen. Lester et al.~\cite{lester2021power} adopt prompt tuning to learn continuous prompt embeddings.
In VLM settings, prompting has similarly been used to align visual representations with textual descriptions. Radford et al.~\cite{radford2021learning} introduce text prompts for zero-shot classification. Building on this idea, Zhou et al.~\cite{zhou2022conditional} propose Context Optimization, which replaces manually designed prompts with a set of learnable vectors.
In this work, we adopt zero-shot prompting in all our experiments and build on the idea of chain-of-thought prompting to encourage explicit reasoning in one of our experiments.
 

\subsection{Datasets}

Several benchmark datasets are commonly used by the sign language research community to evaluate isolated sign language recognition systems, differing in language, vocabulary size, and recording conditions. WLASL~\cite{LI_2020_WACV} and MSASL~\cite{joze2018ms} are among the most widely adopted large-scale ISLR datasets for American Sign Language, featuring thousands of sign instances collected from online videos with diverse signers and visual environments. Other datasets focus on different sign languages and acquisition settings, including NMFs-CSL~\cite{10_1145_3436754}, and SLR500~\cite{8466903} for Chinese Sign Language, as well as AUTSL~\cite{9210578} for Turkish Sign Language. These datasets typically consist of temporally trimmed video clips annotated with a single gloss label and serve as standard benchmarks for ISLR evaluation. 

In this work, we focus on WLASL300 as a widely adopted ISLR benchmark that offers a practical compromise between computational cost and label-set diversity: it is large enough (300 glosses) to meaningfully probe open-vocabulary behavior, while remaining tractable for multi-model, multi-prompt evaluation at scale. To the best of our knowledge, the current state of the art on WLASL300 is DSLNet~\cite{DSLNet}, achieving 89.97\% Top-1 accuracy.

WordNet~\cite{miller-1994-wordnet} provides a lexical database from which we use lists of synonyms. HandSpeak~\cite{handspeak} is an online ASL dictionary from which we utilize descriptions of ASL glosses.

\section{Experiments and Results}

In this section, we first detail our experimental setup in Section~\ref{sec:setup} and then introduce our experiments: 

We benchmark a diverse set of off-the-shelf VLMs under a unified zero-shot ISLR protocol in Section~\ref{sec:zero-shot}. We then analyze prompt sensitivity and task definition through (i) binary sign detection with textual gloss descriptions under different instruction styles, Section~\ref{sec:binary}, and (ii) dataset-knowledge scenarios that progressively constrain the output space from generic ASL to the WLASL300 dataset and, finally, an explicit list of all 300 candidate glosses, Section~\ref{sec:knowledge}.

To account for the broadness of VLMs' possible outputs we add a synonym-aware metric that accounts for synonyms of the ground truth glosses. We assign a list of synonyms to each ground truth gloss that we consider as correct predictions. The synonyms are taken from WordNet~\cite{miller-1994-wordnet}. When two WLASL glosses are synonymous, we treat either as correct.

\subsection{Experimental Setup}
\label{sec:setup}

All open-source VLMs were executed locally on a single NVIDIA H100 GPU using Hugging Face \texttt{transformers}~\cite{transformers} with mixed precision and automatic device placement. Qwen2.5-VL and Qwen3-VL were loaded in \texttt{bfloat16} with FlashAttention-2 and accept a \texttt{video} input plus an explicit sampling rate (FPS) through the Qwen-VL utilities. LLaVA-NeXT-Video was executed with the model’s video processor by extracting a fixed number of frames via PyAV~\cite{PyAV} and passing them as a video tensor to the processor, the “system-like” instruction is embedded directly in the user message text here. InternVL3.5 was deployed as an image-text model by uniformly sampling frames and feeding them as a multi-image chat input; similarly, the system instruction is concatenated with the question inside the user text alongside the image placeholders. BAGEL-7B-MoT was instantiated from its component modules (LLM + SigLIP ViT + VAE) and loaded via \texttt{accelerate}~\cite{accelerate} checkpoint dispatch; inference uses a stack of sampled frames converted to PIL~\cite{pillow} images and passed to \texttt{model.chat}, whose interface exposes a single \texttt{prompt} string rather than separate \texttt{system}/\texttt{user} turns (the script therefore constructs a combined instruction+query prompt when needed).

Proprietary models were queried through official SDKs: GPT-5(-nano) via the OpenAI Chat Completions API with frames encoded in \texttt{base64}, and Gemini via the Google GenAI API with frames sent as JPEG byte parts; therefore, their runtimes reflect end-to-end API latency rather than GPU kernel time.

To ensure comparability across models, all runs use the same baseline prompting protocol; differences across rows in all result tables reflect only the model choice and the visual sampling strategy supported by the corresponding interface (e.g., FPS-based video input vs.\ a fixed number of uniformly sampled frames per clip). This split is purely an interface constraint. 

Qwen2.5-VL and Qwen3-VL expose an explicit \texttt{video} modality where the model (and its processor) handle temporal sampling internally given an FPS setting, so we pass the raw video path and FPS. In contrast, several “video” VLM pipelines (and some APIs) effectively operate as multi-image models: they accept a list of frames/images rather than an encoded video file. For these models we therefore convert each clip into uniformly sampled frames (8/32/64 depending on the experiment) and feed them as a sequence of images; LLaVA-NeXT-Video receives frames through its video processor, while InternVL and BAGEL receive frames as multi-image chat input. Finally, the GPT-5 and Gemini runs in our repository are implemented as frame-based requests (base64 images / JPEG byte parts), so we evaluate them on the same uniformly sampled frames to ensure a consistent comparison under their supported input format.

\subsection{Analysis of zero-shot ISLR with VLMs}~\label{sec:zero-shot}

We begin with a direct zero-shot multi-class recognition setting: each model receives a single trimmed video clip containing one sign and must output a single gloss prediction. 

All models were queried using the same system instruction (with only minor modality wording differences between video vs.\ frame-based interfaces and system prompt placement, see Section~\ref{sec:setup}):
\begin{quote}\small
\texttt{You are an expert ASL (American Sign Language) gloss annotator. Your task is to watch ASL videos and predict the EXACT single English gloss that represents the sign being performed. Provide the gloss of the sign language video, output only the gloss, no other text. Do NOT output phrases like `woman signing' or `hand movements' or descriptions. Examples of correct ASL glosses: BOOK, RUN, HAPPY, WATER, SCHOOL, COMPUTER, DANCE, etc.}
\end{quote}

The user query was kept minimal to make outputs easy to score; for video-capable local models it was:
\begin{quote}\small
\texttt{What is the ASL gloss for this sign? Output only the single English word or phrase.}
\end{quote}

For API-based (frame) models we used the equivalent wording referencing sequential frames:
\begin{quote}\small
\texttt{What is the ASL gloss for this sign shown in these sequential frames? Output only the single English word or phrase.}
\end{quote}

We evaluate zero-shot ISLR on the WLASL300 test split (668 test videos), defining each clip as a single-step multimodal instruction-following problem: given a short ASL clip, the model must output exactly one English gloss (word or short phrase). For each VLM, we run inference with a fixed prompt template and report exact-match Top-1 accuracy and a synonym-aware variant in Table~\ref{tab:wlasl300_accuracy}, together with runtime measured either locally on a single Nvidia H100 GPU or via the corresponding provider API.

We observe that most models perform poorly on the WLASL300 dataset. Unlike traditional classification models, which typically have the number of output neurons equal to the number of possible predictions, VLMs have a much larger distribution of possible outputs, making random guesses significantly less likely to be correct. This is especially true when the model does not know which dataset is used or has no knowledge about the dataset. Some models are more honest (for example, the Nemotron model says 335 times that it does not know), which lowers their measured performance.

\begin{table*}[ht]
\centering
\resizebox{\textwidth}{!}{%
\begin{tabular}{c l l c c c c}
\cmidrule(lr){1-7}
 & \textbf{Model} & \textbf{Input Format} & \textbf{Top-1} & \textbf{Top-1 + Syn.} & \textbf{Time / vid (s)} & \textbf{Total (h)} \\
\cmidrule(lr){1-7}

\multirow{12}{*}{\rotatebox{90}{Open-source}}

& LLaVA-NeXT-Video-7B~\cite{llavanextvideo} & 25 fps & 0.30 & 0.45 & 0.55 & 0.10 \\
\cmidrule(lr){2-7}

& \multirow{2}{*}{InternVL3.5-14B~\cite{internvl3.5}} 
 & 32 frames/clip & 0.30 & 1.20 & 11.81 & 2.19 \\
& & 64 frames/clip & 0.00 & 0.17 & 26.00 & 4.22 \\
\cmidrule(lr){2-7}

& \multirow{2}{*}{Qwen3-VL-30B-A3B-Instruct~\cite{qwen3-vl}} 
 & 4 fps  & \textbf{1.35} & \textbf{1.80} & 0.76 & 0.14 \\
& & 25 fps & 1.20 & 1.35 & 3.22 & 0.60 \\
\cmidrule(lr){2-7}

& \multirow{2}{*}{Qwen3-VL-32B-Instruct~\cite{qwen3-vl}} 
 & 4 fps  & 0.45 & 1.35 & 0.59 & 0.11 \\
& & 25 fps & 0.75 & 1.65 & 2.80 & 0.52 \\
\cmidrule(lr){2-7}

& \multirow{2}{*}{Qwen2.5-VL-7B-Instruct~\cite{qwen25}} 
 & 4 fps  & 0.30 & 0.45 & 0.80 & 0.15 \\
& & 25 fps & 0.60 & 1.20 & 4.60 & 0.85 \\
\cmidrule(lr){2-7}

& \multirow{2}{*}{BAGEL-7B-MoT~\cite{bagel}} 
 & 8 frames/clip  & 0.15 & 0.30 & 3.59 & 0.67 \\
& & 64 frames/clip & 0.15 & 0.30 & 3.92 & 0.73 \\
\cmidrule(lr){2-7}

& Nemotron-Nano-VL-8B-V1~\cite{nvidia2025nvidianemotronnanov2} & 8 frames/clip & 0.15 & 0.60 & 1.30 & 0.24 \\

\midrule
\midrule

\multirow{8}{*}{\rotatebox{90}{Proprietary}}

& \multirow{2}{*}{GPT-5-nano~\cite{gpt-5}} 
 & 8 frames/clip  & 1.20 & 1.20 & 18.59 & 3.45 \\
& & 64 frames/clip & 1.05 & 1.35 & 21.82 & 4.05 \\
\cmidrule(lr){2-7}

& \multirow{3}{*}{GPT-5~\cite{gpt-5}} 
 & 8 frames/clip  & 15.57 & 16.32 & 57.47 & 10.66 \\
& & 32 frames/clip & 13.17 & 13.92 & 72.56 & 14.46 \\
& & 64 frames/clip & 11.98 & 12.87 & 73.36 & 13.61 \\
\cmidrule(lr){2-7}

& Gemini-2.0-flash~\cite{gemini2.5} & 8 frames/clip & 2.65 & 3.48 & 2.22 & 0.37 \\
\cmidrule(lr){2-7}

& \multirow{2}{*}{Gemini-2.5-pro~\cite{gemini2.5}} 
 & 8 frames/clip  & 23.65 & 25.45 & 10.47 & 1.94 \\
& & 64 frames/clip & \textbf{24.40} & \textbf{25.60} & 12.65 & 2.35 \\

\cmidrule(lr){1-7}
\end{tabular}
}
\caption{WLASL300 (668 test split videos): Top-1 accuracy \% (exact match) and synonym-aware Top-1 accuracy, together with average per-video processing time and total runtime, measured in the OpenAI API for GPT-5, the Google AI API for Gemini, and on a single NVIDIA H100 GPU for the remaining models. Reminder: SOTA on WLASL300 with a specialized closed-set classification model is 89.97\%.}
\label{tab:wlasl300_accuracy}
\end{table*}


\subsection{Binary Classification}~\label{sec:binary}

The following experiment is designed to test the capabilities of VLMs that perform poorly in direct multi-class gloss prediction. We assess what ASL knowledge these models exhibit and whether providing additional gloss information improves recognition for models not trained on ASL. We evaluate two strong open-source baselines (Qwen3-VL-32B-Instruct and Qwen3-VL-30B-A3B-Instruct) alongside the best-performing proprietary model in our classification benchmark (Gemini-2.5-Pro). We assess whether the model can identify a target gloss from its textual description, measuring how additional gloss information affects recognition. We provide the model with a description of a gloss and ask whether the described gloss is in the input video. The descriptions used for this experiment are taken from HandSpeak~\cite{handspeak}. Glosses that did not have a description at the time of this experiment are removed in this experiment.

To mitigate models’ tendency to agree with the prompt, we evaluate multiple instruction variants that encourage more critical decision-making. In the \textit{non-skeptical} scenario, the VLM is instructed to describe the video, compare it with a given description, and provide a conclusion. In the \textit{skeptical} scenario, the model is instructed to first explain how the video does not fit the description, then how it does fit, before giving a conclusion. In the scenario with \textit{no description}, only the gloss name is provided in the input, instead of the gloss description, and the VLM is instructed to describe the video and provide a conclusion. We evaluate the final word of the output, which should be '\textit{yes}' or '\textit{no}'. For positive instances, the input video matches the gloss described in the prompt; for negative instances, we sample a video from a different gloss. We balance positives and negatives per gloss description, so a random yes/no predictor has expected precision and recall of 0.5.

The results in Table~\ref{tab:detection_rates} demonstrate the models’ ability to recognize the gloss based on the provided description. Qwen3-VL-32B-Instruct and Qwen3-VL-30B-A3B-Instruct show improved performance in the \textit{nonskeptical} scenario when descriptions are provided compared to the scenario without descriptions. This suggests that these models have limited prior knowledge of the glosses and benefit from explicit textual descriptions. In contrast, Gemini-2.5-pro, exhibits a slight decrease in performance in the \textit{nonskeptical} scenario when descriptions are included. This may be explained by the model’s prior knowledge of the glosses, which could be represented internally in a form that differs from the provided textual descriptions, potentially introducing ambiguity. Introducing the \textit{skeptical} scenario reduced the number of false positive predictions for all models. However, despite this improvement in precision-related behavior, the overall F1 score decreased for all models under the \textit{skeptical} condition.

\begin{table*}[ht!]
\centering
 \begin{tabular}{lcccccc} 
  \hline
  & & \multicolumn{2}{c}{True} & \multicolumn{2}{c}{False}\\
  \textbf{Model} & \textbf{Input} & Positive & Negative & Positive & Negative & F1\\ [0.5ex] 
   \hline
\multirow{3}{*}{Qwen3-VL-30B-A3B-Instruct} & Nonskeptical & 425 & 116 & 380 & 73 & 0.65\\ 
 & Skeptical & 166 & 429 & 69 & 332 & 0.45\\ 
 & No description & 299 & 232 & 263 & 198 & 0.56\\ 
 \hline
\multirow{3}{*}{Qwen3-VL-32B-Instruct} & Nonskeptical & 347 & 335 & 163 & 151 & 0.69\\ 
 & Skeptical & 197 & 430 & 68 & 300 & 0.52\\ 
 & No description & 234 & 312 & 185 & 263 & 0.51\\ 
 \hline
\multirow{3}{*}{Gemini-2.5-pro} & Nonskeptical & 339 & 456 & 42 & 159 & 0.77\\ 
 & Skeptical & 250 & 464 & 34 & 248 & 0.64\\ 
 & No description & 470 & 262 & 236 & 28 & 0.78\\ 
 \hline
\end{tabular}
\caption{\textbf{Comparison of instruction variants for binary sign detection}, where the model judges whether an input video matches a provided gloss description. \textit{Nonskeptical}: the model describes the video, compares it to the description, and decides. \textit{Skeptical}: the model first argues why the video does \textbf{not} match, then why it \textbf{does}, and finally decides. \textit{No description}: the model is given only the gloss name (no textual description).}
\label{tab:detection_rates}
\end{table*}

\subsection{Dataset Knowledge}~\label{sec:knowledge}

In the following experiment, we examine how explicit knowledge of the WLASL300 label space affects model performance. We compare prompts with increasing specificity: (i) stating only that the inputs are ASL videos, (ii) additionally stating that the videos come from WLASL300, and (iii) listing all 300 candidate glosses in the prompt. As shown in Table~\ref{tab:different_knowlege_prompts}, merely naming the dataset provides little to no benefit, whereas enumerating the candidate glosses substantially improves accuracy and increases the fraction of predictions that fall within the WLASL300 label set. Notably, the relative gain from synonym-aware scoring is smaller when the full gloss list is provided.

\begin{table*}[ht!]
\centering
\begin{tabular}{l c c c c}
\hline
\textbf{Model} & \textbf{Knowledge} & \textbf{Top-1} &  \textbf{Top-1 + Syn.} & \textbf{PID}\\
\hline
\multirow{3}{*}{Qwen3-VL-30B-A3B-Instruct}&It's ASL& 1.20 & 1.35 & 56.59\\ 
&It's WLASL300& 2.10 & 2.10 & 48.35\\
&All glosses listed& 2.25 & 2.54 & 88.92\\
\hline
\multirow{3}{*}{Qwen3-VL-32B-Instruct}&It's ASL& 0.45 & 1.20 & 16.47\\ 
&It's WLASL300& 0.45 & 1.20 & 27.84\\
&All glosses listed& 1.50 & 1.80 & 87.13\\
\hline
\multirow{3}{*}{Gemini-2.5-pro}&It's ASL& 23.93 & 25.15 & 52.90\\
&It's WLASL300& 23.80 & 25.30 & 50.00\\
&All glosses listed & 32.28 & 33.33 & 84.38\\
\hline
\end{tabular}
\caption{\textbf{Comparison of dataset-knowledge prompting scenarios}. \textit{It’s ASL}: the prompt specifies only that the inputs are ASL videos. \textit{It’s WLASL300}: the prompt additionally states that the videos come from the WLASL300 dataset. \textit{All glosses listed}: the prompt includes the full list of 300 WLASL300 candidate glosses. Results are reported for Qwen3-VL-30B-A3B-Instruct (4 fps), Qwen3-VL-32B-Instruct (4 fps) and Gemini-2.5-pro (8 frames/clip), including Top-1 accuracy, synonym-aware Top-1 accuracy, and PID (Predictions in Distribution), which denotes the percentage of predictions that fall within the WLASL300 label set (i.e., the predicted gloss is in-distribution).}
\label{tab:different_knowlege_prompts}
\end{table*}

When the full gloss list is included, Qwen3-VL-32B-Instruct exhibits a strong position bias, predicting glosses that appear earlier in the prompt more frequently than those appearing later. With an alphabetically sorted list, the most common prediction is “ABOUT” (457 videos), the first entry; when the list order is reversed, the most frequent predictions shift to “YOUR” (119 videos) and “YOU” (96 videos), which become the final entries. This bias is also prevalent in Qwen3-VL-30B-A3B-Instruct. Gemini-2.5-pro does not show a comparable ordering effect.




\section{Discussion}
Our results show that the current \textbf{publicly available open-source models} are not suitable for the zero-shot ISLR task yet. The best-performing model, Qwen3-VL, achieves performance substantially below the current state-of-the-art. However, the experiments show a strong understanding of visual scenes and some knowledge of handshapes, hand movements, and facial expressions. We believe that this knowledge was obtained during training from other task domains, and Qwen3-VL is able to generalize to this new, unseen task. 

On the other hand, proprietary models are able to provide much better results overall. We argue that the reason for it is twofold. First, proprietary models are likely significantly larger and trained with substantially more compute than the tested publicly available models, which gives them much better generalization ability. Second, their training datasets may include sign language content, which could provide additional useful visual representations. However, due to the lack of transparency regarding the training data of these models, it is not possible to determine whether datasets such as WLASL300 were included.

To conclude, our results suggest that while zero-shot ISLR remains challenging for current VLMs, these models already possess useful visual and semantic representations that can transfer to sign language understanding. This opens several promising research directions, including the integration of structured output constraints, improved prompt design, and lightweight fine-tuning strategies that leverage the strong generalization capabilities of modern multimodal models.

\section{Conclusion}
In this work, we evaluated the capability of modern Vision–Language Models (VLMs) to perform isolated sign language recognition (ISLR) in a zero-shot setting. Our experiments on the WLASL300 benchmark show that current open-source VLMs perform poorly on direct multi-class sign recognition when used purely with prompt-based inference. However, further analyses reveal that these models possess partial visual–semantic understanding of sign language, as demonstrated by stronger performance in binary matching tasks and when the output space is constrained.

Proprietary models achieve notably higher accuracy, suggesting that model scale, training data diversity, and pretraining strategies play an important role in enabling multimodal sign understanding. Overall, our results indicate that while zero-shot ISLR remains challenging, modern VLMs already exhibit promising capabilities that could be leveraged through improved prompting, structured output constraints, or lightweight fine-tuning.

\section*{Acknowledgements}

The work has been supported by the grant of the University of West Bohemia, project No. SGS-2025-011.

Computational resources were provided by the e-INFRA CZ project (ID:90254), supported by the Ministry of Education, Youth and Sports of the Czech Republic.

\bibliographystyle{ieeenat_fullname}
\bibliography{bibliography}

\end{document}